\documentclass[conference]{IEEEtran}
\IEEEoverridecommandlockouts
\usepackage{cite}
\usepackage{amsmath,amssymb,amsfonts}
\usepackage{algorithmic}
\usepackage{graphicx}
\usepackage{textcomp}
\usepackage{xcolor}
\def\BibTeX{{\rm B\kern-.05em{\sc i\kern-.025em b}\kern-.08em
    T\kern-.1667em\lower.7ex\hbox{E}\kern-.125emX}}
\begin{document}

\title{Robust Traffic Light Detection \\ Using Salience-Sensitive Loss: \\Computational Framework and Evaluations}

\author{\IEEEauthorblockN{Ross Greer* \thanks{*All authors are associated with the Laboratory for Intelligent \& Safe Automobiles (LISA) at the University of California, San Diego in La Jolla, USA.}}
\IEEEauthorblockA{
regreer@ucsd.edu}
\and
\IEEEauthorblockN{Akshay Gopalkrishnan}
\IEEEauthorblockA{
agopalkr@ucsd.edu}
\and
\IEEEauthorblockN{Jacob Landgren}
\IEEEauthorblockA{
jlandgre@ucsd.edu}
\and
\IEEEauthorblockN{Lulua Rakla}
\IEEEauthorblockA{
lrakla@ucsd.edu}
\and
\IEEEauthorblockN{Anish Gopalan}
\IEEEauthorblockA{
agopalan@ucsd.edu}
\and
\IEEEauthorblockN{Mohan Trivedi}
\IEEEauthorblockA{
mtrivedi@ucsd.edu}
}

\maketitle

\begin{abstract}
One of the most important tasks for ensuring safe autonomous driving systems is accurately detecting road traffic lights and accurately determining how they impact the driver's actions. In various real-world driving situations, a scene may have numerous traffic lights with varying levels of relevance to the driver, and thus, distinguishing and detecting the lights that are relevant to the driver and influence the driver's actions is a critical safety task.
This paper proposes a traffic light detection model which focuses on this task by first defining salient lights as the lights that affect the driver's future decisions. We then use this salience property to construct the LAVA Salient Lights Dataset, the first US traffic light dataset with an annotated salience property. Subsequently, we train a Deformable DETR object detection transformer model using Salience-Sensitive Focal Loss to emphasize stronger performance on salient traffic lights, showing that a model trained with this loss function has stronger recall than one trained without.
\end{abstract}


\section{Introduction}

\subsection{Overview}

Accurate detection and recognition of traffic lights is vital for an autonomous vehicle to observe and interact with its surroundings in a safe manner, communicating information relevant to predicting an agent's trajectory \cite{deo2020trajectory} \cite{messaoud2021trajectory} \cite{greer2021trajectory} \cite{mogelmose2015trajectory} or enabling ADAS features \cite{greer2023champ}. In general, standard object detectors for traffic lights, signs, or pedestrians operated by proposing regions of interest, which involves considering a standard set of anchors or window centers within an image, and classifying the contents of the found region \cite{philipsen2015traffic} \cite{mogelmose2015detection}. However, approaches involving regions of interest are generally limited by the computational costs and gaps in coverage that come with creating convolutional filters that cover an entire image. As an alternative, the transformer model has been proposed to cover an entire image and select only features that are relevant to the region of interest. Transformers allow for more parallelization than older approaches and thus, reduce training times. To minimize the problem of computational costs that the regions of interest approach has, the transformer approach has been further refined to include a stage of learning (via a limited number of deformable attention heads) where an image should be sampled to extract meaningful relational features to the region of interest. As defined in \cite{greer2023salient}, this approach is known as the Deformable Detection Transformer, described in technical detail in the following section.

While this transformer-based approach may improve traffic-light detection, consider a driving scene with numerous traffic lights that are simultaneously presented to an autonomous driving system, containing some lights that are relevant to the vehicle and some of that are not. Ideally, a detector will be able to precisely recall the statuses of all traffic lights in an image and make complex driving-related decisions accordingly. However, factors such as environmental noise and underrepresented driving situations may lead to errors in detectors, but some errors may be preferred over others. For instance, it is less critical that a vehicle making a right turn while in the right lane sees the status of the traffic light corresponding to the left lane, or that a vehicle continuing straight along an intersection sees the status of a traffic light perpendicular to the car's current trajectory. We ascribe this quality of pertinence and attention-worthiness to the word salience, as introduced in the context of traffic signs in \cite{greer2022salience}, adopting an equivalent definition in the following section. We illustrate the use of our salience-sensitive computational framework in Figure 1. 
\begin{figure*}
    \centering
    \includegraphics[width=\textwidth]{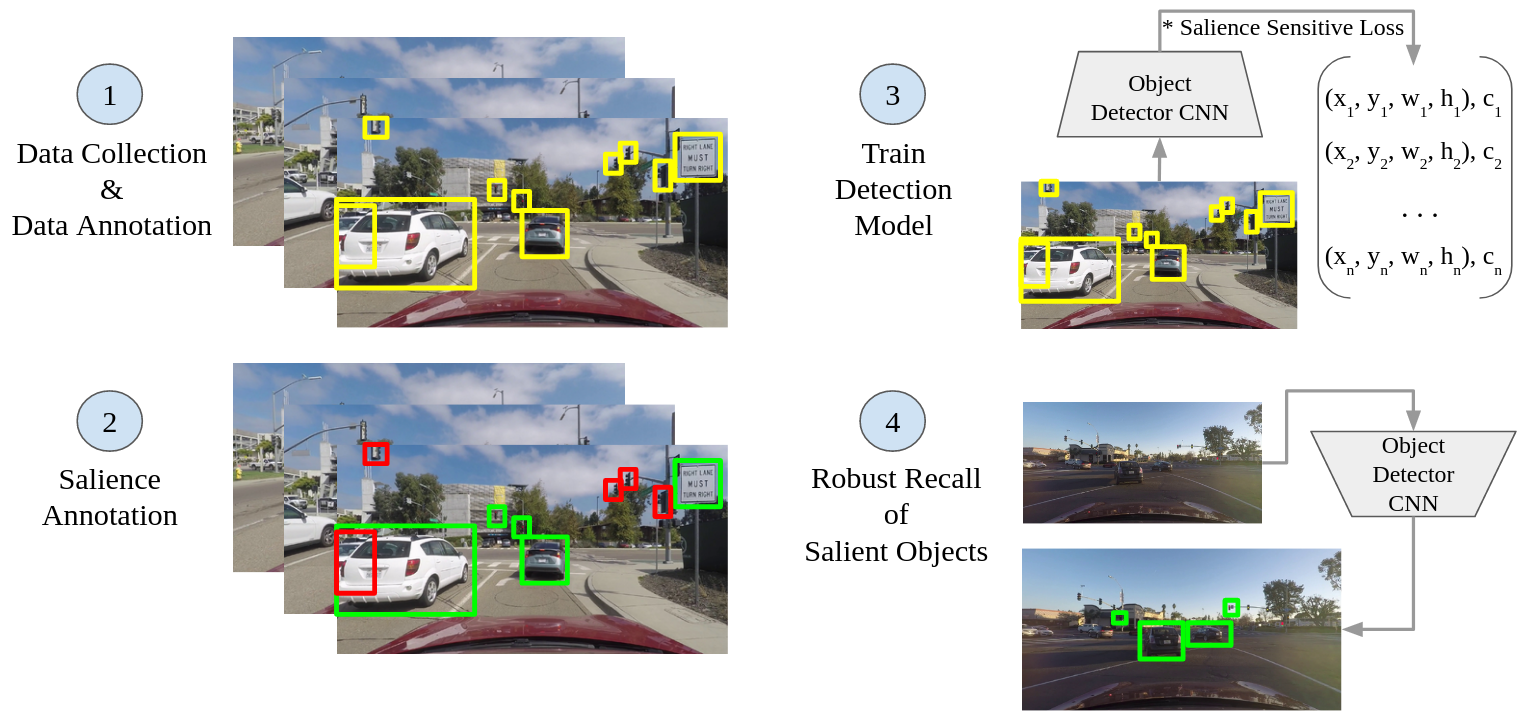}
    \label{salienceflow}
    \caption{The general computational framework for leveraging object salience in deep learning tasks. First, data is collected and annotated per task specifications. The data is provided an additional annotation, salience, which can be provided by either an expert annotator or computer model. The salience property is utilized during the training process, leading to a robust recall of salient objects during deployment inference.}
\end{figure*}

\subsection{Salience}

\begin{figure}[htb!]
    \centering
    \includegraphics[width=.35\textwidth,height=.13\textheight]{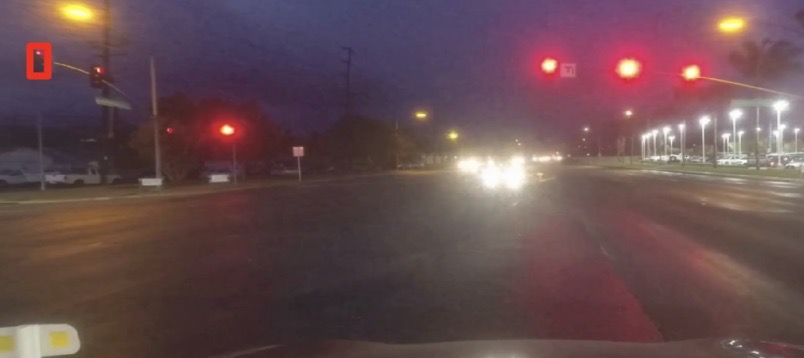}

    \includegraphics[width=.35\textwidth,height=.13\textheight]{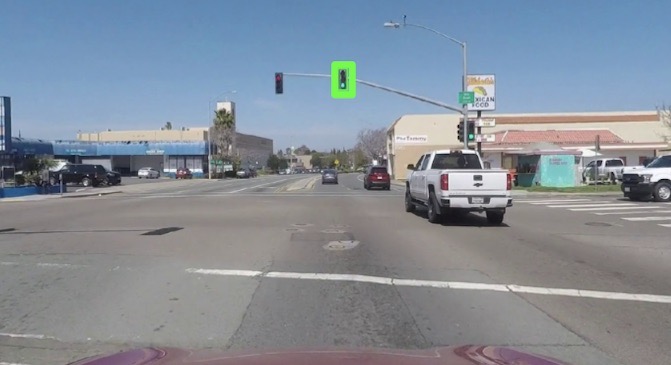}
    \includegraphics[width=.35\textwidth,height=.13\textheight]{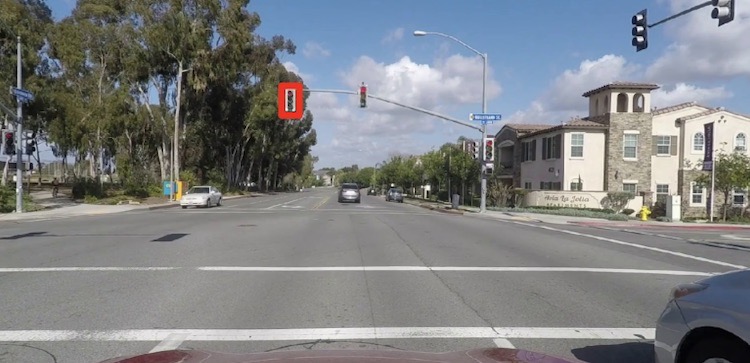}
    
    \caption{Annotated examples of the \textit{salience} property for selected traffic light scenarios. A red box indicates a non-salient example while a green box indicates a salient example. Since the traffic light in the top image is in a different intersection than the driver, the light is not considered to be salient. In the middle image, the traffic light is in the same intersection of the driver and indicates the direction the driver is going in so it is considered to be salient. Finally, the bottom image is not salient since the driver is not in a left turn lane and thus the left turn traffic light is not relevant to the driver.}
    \label{fig:1}
\end{figure}

As defined in \cite{greer2022salience}, a traffic light is salient if it directly influences the next immediate decision to be made by the ego vehicle if no other vehicles were present on the road. For instance, consider if a car were to be driving straight through an intersection. We would classify the vehicle-facing straight traffic light as salient, as its status will influence whether the vehicle will stop or proceed. However, we would not classify the protected traffic lights for left and right turns as salient, as the status of these lights is irrelevant to the vehicle's decision. In general, after taking into account factors such as the ego vehicle's current lane and next vehicle maneuver, we would selectively classify traffic lights as salient. Similar to the definition in \cite{greer2022salience} involving traffic signs, in the case of multiple sequential intersections or traffic lights that are present in the same frame of data, only the traffic lights that are relevant to the vehicle's immediate next action would be important to detect, and thus, would be classified as salient. For example, if the ego-vehicle is in a left-turn lane, then we would classify the traffic light signaling the left-turn as salient, while the non left-turn traffic lights in the same intersection would be considered non-salient.  Figure \ref{fig:1} shows qualititative examples of salient and non-salient traffic light annotations from our collected dataset.

We propose that salience-aware training methods can be used to improve the training of traffic light detection systems. We make two contributions: (1) creation of the first salience-annotated LAVA Salient Lights Dataset, and (2) using the concept of Salience-Sensitive Focus Loss defined in \cite{greer2023salient}, evaluation of the impact of Salience-Sensitive Focal Loss while training detection transformer models.

\section{Related Research}

\subsection{Traffic Light Detection and Classification Models}

There are various issues and challenges related to traffic light detection.
\begin{enumerate}
    \item Illumination: Images containing traffic lights may contain different illumination due to various environmental factors \cite{gautam2023image}. Traffic lights also have variability in their own lighting, as sometimes the traffic lights themselves may be off or in different cases (green, yellow, red). 
    \item In driving scenarios such as intersections, traffic lights may appear in a variety of orientations, requiring a robust model that can recognize a traffic light in all different positions.
\end{enumerate}
To specifically address illumination problems for traffic light detection, Shi et al. \cite{shi2015real} proposed a model that is robust to different illumination conditions. The first step of the model uses an adaptive background suppression algorithm to highlight detected traffic lights, and the second step, the recognition module, verifies each candidate regions and classifies the traffic lights. Instead of using hand-designed features or algorithms, deep learning approaches with an abundance of data will allow a model like A CNN do learn filters and features that help identify traffic lights.  Behrendt et al. \cite{behrendt2017deep}  use a YOLO architecture with the classification network removed to perform traffic light detection. They emphasize that this network is optimized for automated driving, as the network is efficient enough to make real-time predictions. Image segmentation can be also be used to detect the locations of traffic lights. Weber et al. \cite{weber2016deeptlr} devise the network DeepTLR, a single deep CNN that outputs a probability map that represents a traffic light being present in a certain region. With this region-wise classification, a regression module is used to predict a set of bounding boxes for detected traffic lights. Ennahhal et a. \cite{ennahhal2019real} evaluate traffic light detection performance on various state-of-the-art object detection models. They find that Faster R-CNN gives the best mean average precision for this task.   

\subsection{Traffic Light Datasets}

Traffic light datasets are invaluable in the realm of autonomous vehicles for building robust traffic light detection and classification systems. There are various traffic light datasets, many of which define various features and categories for traffic lights. The typical categories these labels touch on are: on/off, color, or go/warning/stop. Examples of such public datasets are the Bosch Small Traffic Light Database \cite{behrendt2017deep} and the VIVA challenge dataset \cite{jensen2016vision}. While these datasets have been historically useful for the purposes of traffic light detection and classification, for the purpose of this paper, we want to consider datasets that incorporate the added annotation category of "salience". 

Using this filter, we identified the DriveU Traffic Light Dataset (DLTD) \cite{fregin2018driveu}. This dataset includes 232,039 annotations from 11 different cities in Germany, with each annotation having tags to describe: color, directionality, number of lamps, orientation, occlusion level, and any other visual abnormality. In addition to these tags, the DLTD dataset also includes a special feature property tag they name "relevance", which they define to correspond to traffic lights that transport the information relevant to the planned route of the vehicle. This definition for ``relevance" is equivalent to our definition for ``salience". Our proposed dataset, the LAVA Salient Lights Dataset, shares this same key attribute of relevance/salience, and for the first time applies salience to US traffic lights. 

\begin{table*}
\caption{Information on two traffic light datasets which are annotated with the salience feature. The LAVA Salient Lights is the first to include traffic lights from the United States.}
\label{Dataset-table}
\vskip 0.15in
\begin{center}
\begin{tabular}{||c c c c ||}
\hline
Dataset & Num. Of Images & Country & Important Features \\
\hline\hline %[0.5ex]
 DriveU Traffic Light Dataset (DLTD) & 232,039 & Germany & color, directionality, num. of lamps, orientation, occlusion, ``relevance"  \\
\hline
LAVA Salient Lights Dataset & 30,566 & United States & color, directionality, occlusion, ``salience" \\
\hline
\end{tabular}
\end{center}
\end{table*}

\subsection{Traffic Object Salience Research}

The idea of saliency with regard to vehicle objects and visual attention mechanisms has been studied by many researchers. However, the definition of what constitutes a "salient" traffic object is not universally agreed upon. From the literature, we have identified two main categories of object saliency: attentive salience and instructive salience. Attentive salience corresponds to objects or regions drivers tend to look at, despite whether or not they are important to the driver's trajectory and/or decision-making. For attentive salience, a common approach is to monitor the driver's gaze and estimate what objects or regions they are looking at \cite{ohn2015surveillance}. Tawari and Trivedi \cite{tawari2014robust} take such an approach, where driver pose dynamic information was used to estimate a driver gaze zone. This system detects and tracks points of interest on the face such as eye corners, nose tips, and nose corners to estimate the head pose and thus predict the driver's gaze. They found that this approach increased performance over using static features such as head pose angles. For an attentive salience system to be robust, it must be invariant to different scales, perspectives, and subjects. To reach this gaze generalization, Vora et al. \cite{vora2017generalizing} utilize a convolutional neural network trained on ten different subjects to estimate the gaze direction. Also instrumental to the development of attentive salience, Dua et al. \cite{dua2020dgaze} construct the first large-scale driver gaze mapping dataset, DGAZE, which allows for further analysis of driver gaze in different road and traffic conditions. The SAGE-Net model developed by Pal et al. \cite{pal2020looking} learns attentive salience using attention mechanisms to predict an autonomous vehicle's focus of attention. This model utilizes driver gaze alongside other important metrics such as the distance to objects and the ego-vehicle's speed to evaluate object saliency. Tawari et al. \cite{tawari2018learning}, on the other hand, represent driver gaze behavior for a sequence of frames by building a saliency map via a fully convolutional RNN. This map uses three pixel classes: salient pixels, non-salient pixels, and neutral pixels. Attentive salience is also important for predicting driver maneuvers and braking intent. Ohn-Bar et al. \cite{ohn2015surveillance} utilize a head pose prediction model to predict overtaking and braking intent. The system they outline has a strong emphasis on real-time performance, which is extremely important for attentive salience models.

Instructive salience, on the other hand, aims to prioritize traffic objects that are critical to the ego-vehicle's future trajectory. This version of salience emphasizes objects/regions that a driver should be observing in order to maneuver the vehicle properly, and is the version of salience that our work focuses on in regards to traffic lights. Instructive salience models are typically more costly than attentive models due to the manual labeling of important objects with respect to the ego-vehicle. This type of labeling is more cognitively demanding as the annotator must take into consideration both the ego-vehicles current position and future trajectory. To avoid this manual step, Bertasius et al. \cite{bertasius2017unsupervised} utilize an unsupervised learning approach for detecting important objects without any saliency labels. This unsupervised model uses a segmentation network that proposes possible important objects, which are then fed to a recognition agent that incorporates spatial features to predict the important objects. Instead, using a supervised learning process, Greer et al. \cite{greer2022salience} classify the saliency of road signs, which can be utilized for efficient dataset annotation. This led to the LAVA Salient Signs Dataset \cite{greer2023salient}, a dataset of 31,191 labeled traffic signs with a validated salience property. Lateef et al. \cite{lateef2021saliency} utilize a conditional GAN that predicts the important things a driver should be looking at within a traffic scene, matching the definition we use for instructive salience. For this model, they incorporate semantic labels from existing datasets and use saliency detection algorithms to predict which traffic objects are most important. Zhang et al. \cite{zhang2020interaction} also create a model for instructive saliency using interaction graphs that estimate object importance in driving scenes. They note that their object important definition corresponds to the objects that help with the driver's real-time decision-making, falling in line with our definition for instructive salience.

\section{Methods}
\subsection{Data Collection}
The LISA Amazon-MLSL Vehicle Attributes (LAVA) dataset contains labeled bounding boxes of traffic lights taken from a front-facing camera of a vehicle. This dataset was collected from the greater San Diego Area and contains a variety of road types, lighting, and traffic conditions. The traffic lights can be categorized with a status of on, off, or undefined, with a color of red, yellow, green, or undefined (traffic light is not on), true/false attributes for whether the light is directional or not, and with an occlusion level of non-occluded, partial, or major. Additionally, we added an additional light salience property for all traffic light annotations that were performed as either true or false. Due to the potentially ambiguous nature of saliency with regard to any given traffic snapshot, this additional property brings added difficulty with respect to accurate data annotation. As there is some level of subjectivity to a definition for saliency which takes into account a driver's attention or perceived relevance of road objects, it is important that any provided definition is appropriately constrained and that annotations are validated across raters to ensure consistency. A light salience validation process was performed where the boolean salience property for each annotation was double-checked for consistency with our definition of salience. This data collection and validation procedure ensured that we had properly labeled every traffic light as salient or non-salient depending on the traffic scene in the image. The resulting dataset after salience validation is known as the LAVA Salient Lights Dataset. In the process of annotation, the aforementioned definition of salience was used as a standard for annotation, with select frequent ambiguous cases handled according.
to the additional criteria below:
\begin{itemize}
    \item  In the case of a frame where the lane the ego-vehicle is in is indeterminate, both traffic lights pertaining to straight and left turns were labeled salient.
    \item In the case of the ego-vehicle approaching an intersection in the left-most forward-bound lane with an approaching left turn lane opening, both traffic lights pertaining to straight and left turns were labeled salient until it is clear which direction the ego-vehicle will go in.
\end{itemize}
We collected 30,566 traffic light annotations, with 9,051 salient annotations and 21,515 non-salient annotations. Figure 3 shows the frequency of each light type in the  LAVA Salient Lights Dataset.

\begin{figure}
    \centering
    \includegraphics[width=0.49\textwidth]{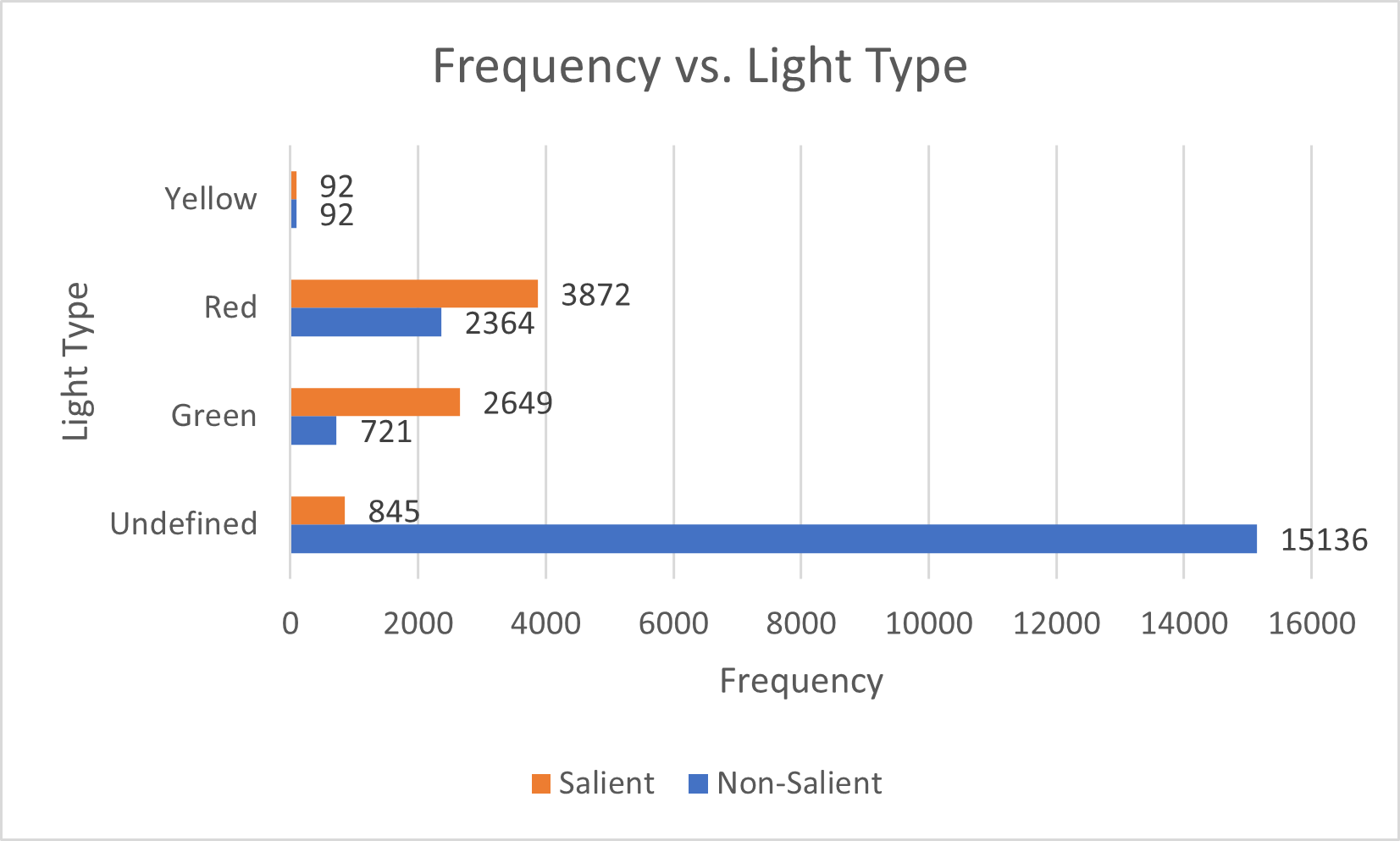}
    \caption{Chart of Light Type Frequencies in the LAVA Salient Lights Dataset. The orange rows represent salient lights and the blue row are for non-salient lights. The undefined non-salient light is the most common light type in the dataset, demonstrating just how many non-salient examples that are present in driving scenarios.}
    \label{fig1}
\end{figure}

\subsection{Light Detection with Deformable DETR}
To perform Traffic Light Detection, we use the end-to-end transformer object detection model Deformable DETR \cite{zhu2020deformable}. For the Deformable DETR hyperparameters, we use a ResNet-50 backbone, 300 attention heads, 50 epochs, a learning rate of 0.0002, and a batch size of 2 due to GPU constraints. We also employ gradient clipping and learning rate decay. We define two approaches to perform the Traffic Light Detection:
\begin{enumerate}
    \item A standard Deformable DETR model trained for traffic light detection 
    \item Deformable DETR model trained with a custom salient-light loss function. 
\end{enumerate}
We compare the two approaches to see if annotating salient traffic lights and focusing performance on such light types increases our detection precision and recall. 
\subsection{Salient-Sensitive Loss for Traffic Lights}
The Deformable DETR involves two steps: bounding box regression via a 3 layer Feed-Forward-Network and bounding box binary classification. For this last step, bounding boxes are classified into either two categories: object or foreground. The classification step is trained through a focal loss equation which emphasizes performance on difficult examples. The focal loss equation is:
\begin{equation}
    FL(p_t) = -\alpha_{FL}(1-p_t)^\gamma \log(p_t)
\end{equation}
This focal loss equation adds a $(1-p_t)^\gamma$ factor to the standard cross-entropy loss function, where $p_t$ represents the probability of a ground truth glass. The $\gamma$ parameter is a focusing parameter which if increased puts more emphasis on more difficult and misclassified examples. $\alpha$ is a hyperparameter is used to balance emphasis on focal loss. We borrow from this focal loss function and customize it such that our loss function emphasizes performance on salient traffic lights. The salient-loss function is defined below:
\begin{equation}
    FL(p_t) = -\alpha_{FL}\omega_{SL}(1-p_t)^\gamma \log(p_t)
\end{equation}
The equation is similar to focal loss equation except for the addition of $\omega_{SL}$. If the nearest ground truth light detection is salient, we set $\omega_{SL} > 1$ such that the loss function is now stricter and requires greater performance on salient lights. We found that the best value for $\omega_{SL}$ on salient examples was 4.  Otherwise if the nearest ground truth detection is non-salient, $\omega_{SL} = 1$ and essentially regular focal loss. In considering system limitations, while in general this loss does not combat the learning of the generic focal less, but rather enhances its effects for important objects. However, one relevant consideration is the care placed in defining what is ``important". Using this salient-sensitive loss function over the traditional focal loss function can open risks if the saliency property within the dataset is inaccurately or inconsistently annotated, hence why it is important to validate salience annotations as previously mentioned. 

\section{Experimental Evaluation}

We evaluate the Deformable DETR Traffic Light Detection Models trained on the LAVA Salient Lights Dataset with and without salient loss. We divided the data at random, with 80\% in training, 10\% in validation, and 10\% in test sets. We trained each model for 50 epochs.

With each detection made by the model, there is a simultaneous confidence score output. We threshold our detections by this confidence score, sweeping across values from 0 to 1 in increments of 0.1. Once we have collected these model predictions, we use IOU with a constant threshold and the ground-truth light detections to calculate the number of correct predictions. Using these sweeping confidence thresholds for detections and using IOU to find the correct predictions, we can generate a precision-recall graph. These charts are shown in Figures \ref{fig1} to \ref{fig3}.

How well do these models recall traffic lights under this training regimen? Deformable DETR appears to be effective at learning to detect traffic lights as a base model, and Figure \ref{fig1} shows that for salient lights, the salience-sensitive loss increases the performance at all points on the precision-recall curve, even giving a stronger concavity indicating sustained recall even under tighter precision thresholds. Similarly, Figure \ref{fig2} also shows a clear separation and concavity for the model trained with salience-sensitive loss. 

Does salience-sensitive loss succeed in prioritizing performance on salient lights versus lights which may be less important to the ego vehicle? Figure \ref{fig3} shows that at high confidence values, the salience-sensitive loss creates a fairly strong difference in performance on salient lights versus all lights in the scene. This means that as the model becomes more scrutinous with an increased confidence threshold, the salient lights remain well-detected compared to the total collection of lights. Qualitative results illustrating model performance are provided in Figure \ref{fig:qual}. 

%At maximal recall, the model trained without salience-sensitive loss recalls 71.77\% of the traffic lights in the test set, while the model trained \textbf{with} salience-sensitive loss recalls 77.03\% of the traffic lights in the test set -- a 12\% increase! Salience-sensitive loss increases recall of specifically the salient lights from 71.98\% to 76.21\%, and of the non-salient lights from 71.68\% to 77.34\%. 

\begin{figure}
    \centering
    \includegraphics[width=0.49\textwidth]{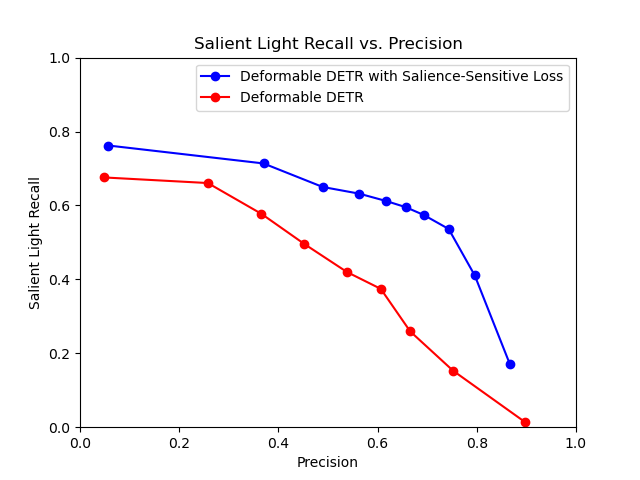}
    \caption{Recall of Salient Traffic Lights versus Precision on All Traffic Lights. The model trained with salience-sensitive loss (in blue) outperforms the model trained without, reaching consistently higher values of recall for similar values of precision.}
    \label{fig1}
\end{figure}

\begin{figure}
    \centering
    \includegraphics[width=0.49\textwidth]{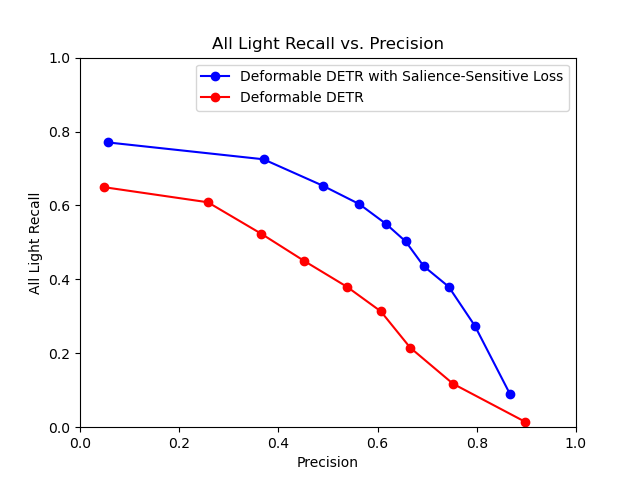}
    \caption{Recall of All Traffic Lights versus Precision on All Traffic Lights. The model trained with salience-sensitive loss (in blue) outperforms the model trained without, reaching consistently higher values of recall for similar values of precision.}
    \label{fig2}
\end{figure}

\begin{figure}
    \centering
    \includegraphics[width=0.49\textwidth]{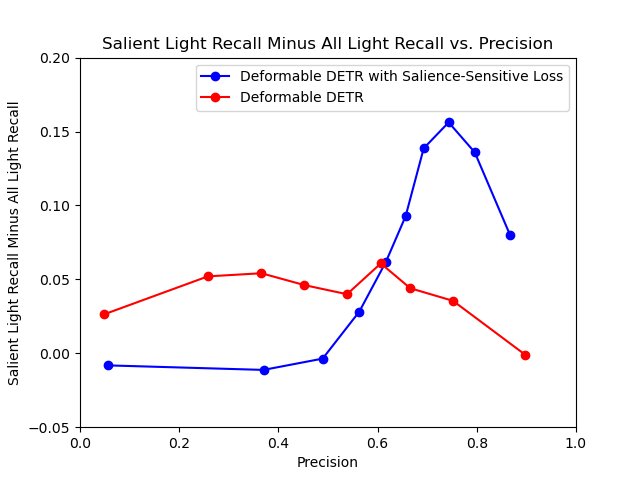}
    \caption{The difference in recall on salient lights versus all lights, plotted against precision on all traffic lights. This graph is meant to address the question \textit{Does salience-sensitive loss successfully prioritize salient traffic lights?} Any time the graph is positive, the model is giving stronger recall of salient lights than overall recall. We see that this occurs naturally (red) without salience-sensitive loss, but by adding salience-sensitive loss (blue), higher confidence thresholds lead to recall difference of greater than 5\%.}
    \label{fig3}
\end{figure}

\begin{figure*}
    \centering
    \includegraphics[width=.48\textwidth]{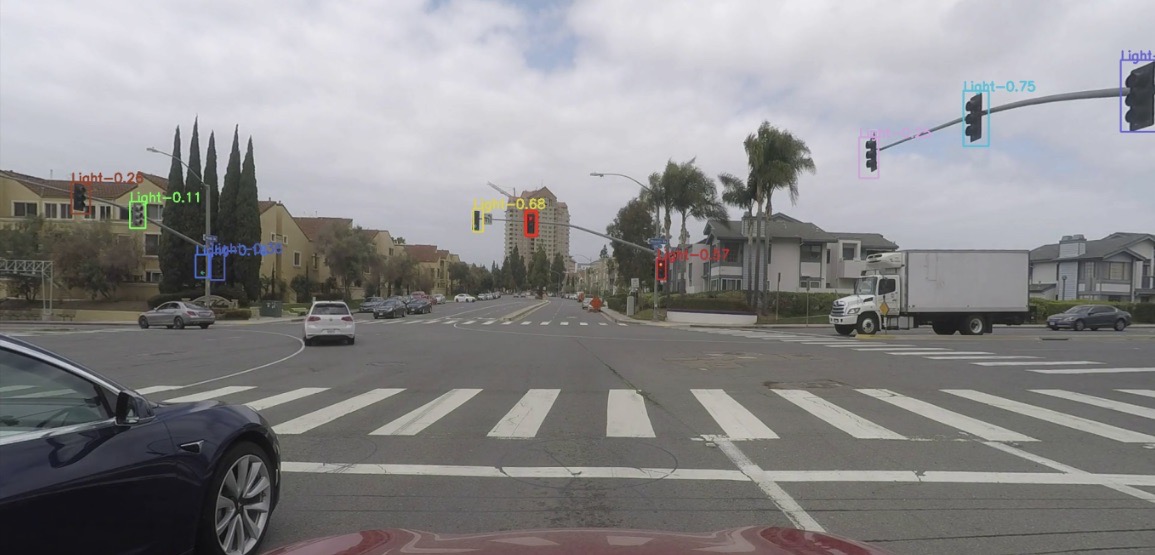}
    \includegraphics[width=.48\textwidth]{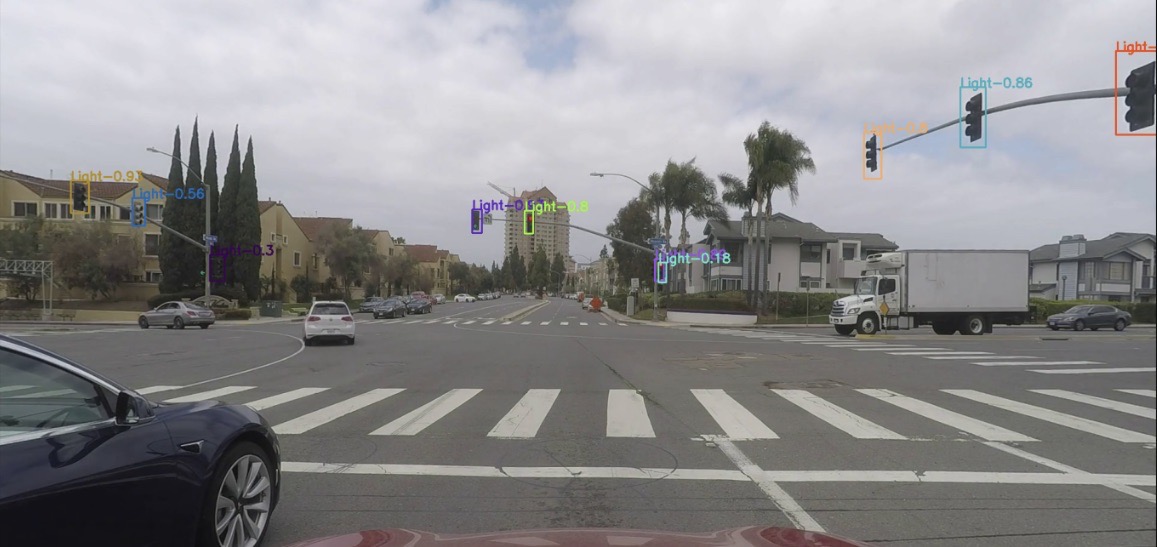}
    \includegraphics[width=.48\textwidth]{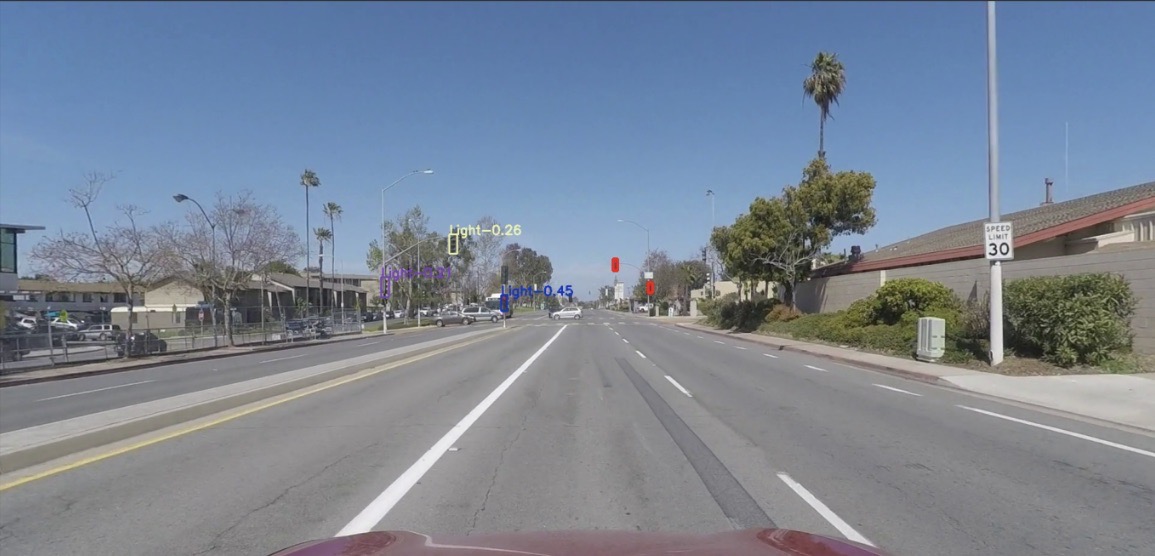}
    \includegraphics[width=.48\textwidth]{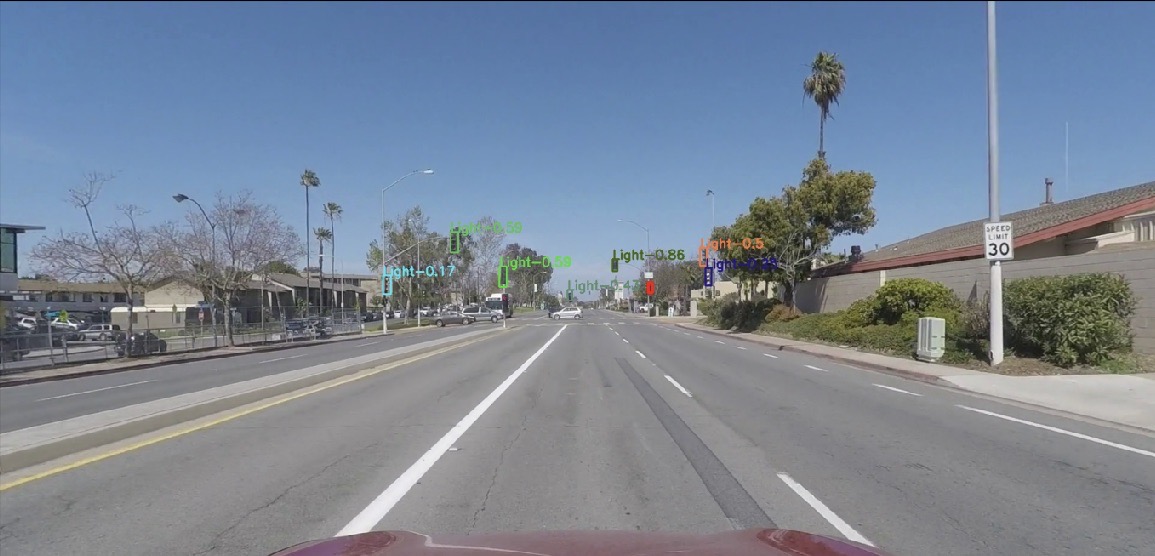}
    \includegraphics[width=.48\textwidth]{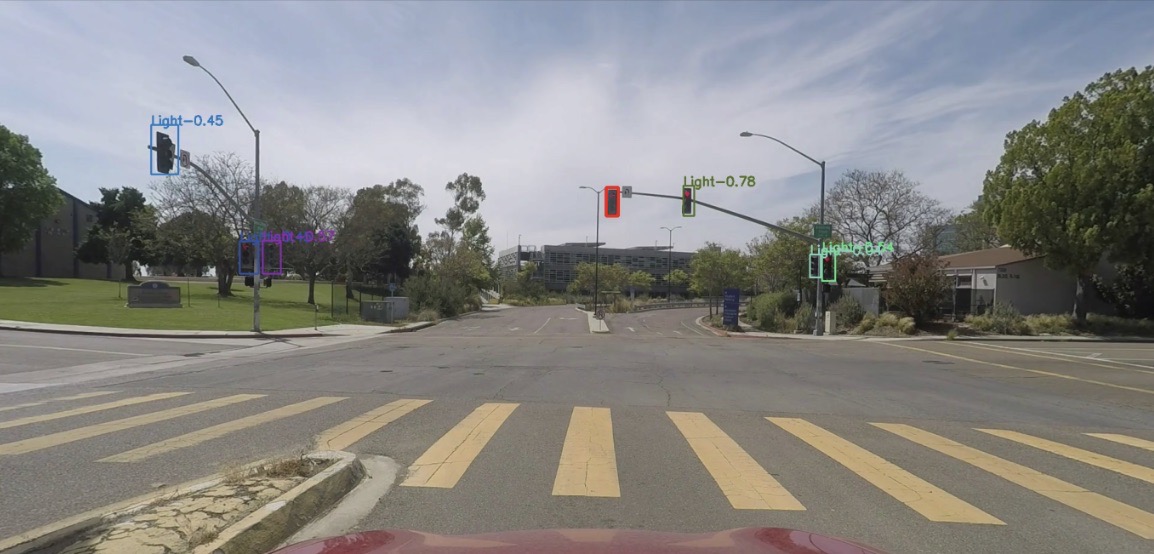}
    \includegraphics[width=.48\textwidth]{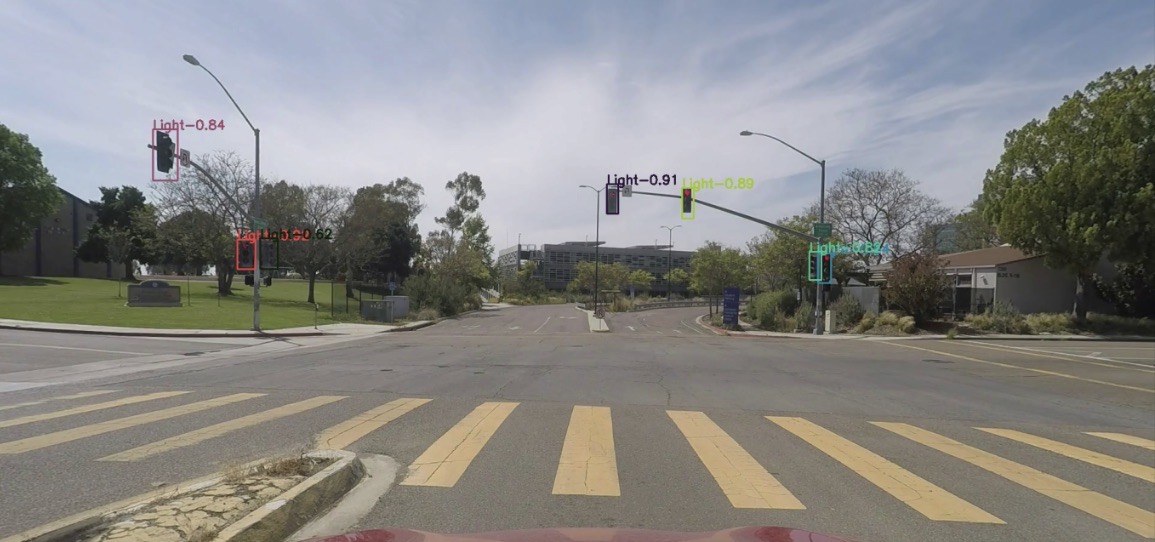}
    \includegraphics[width=.48\textwidth]{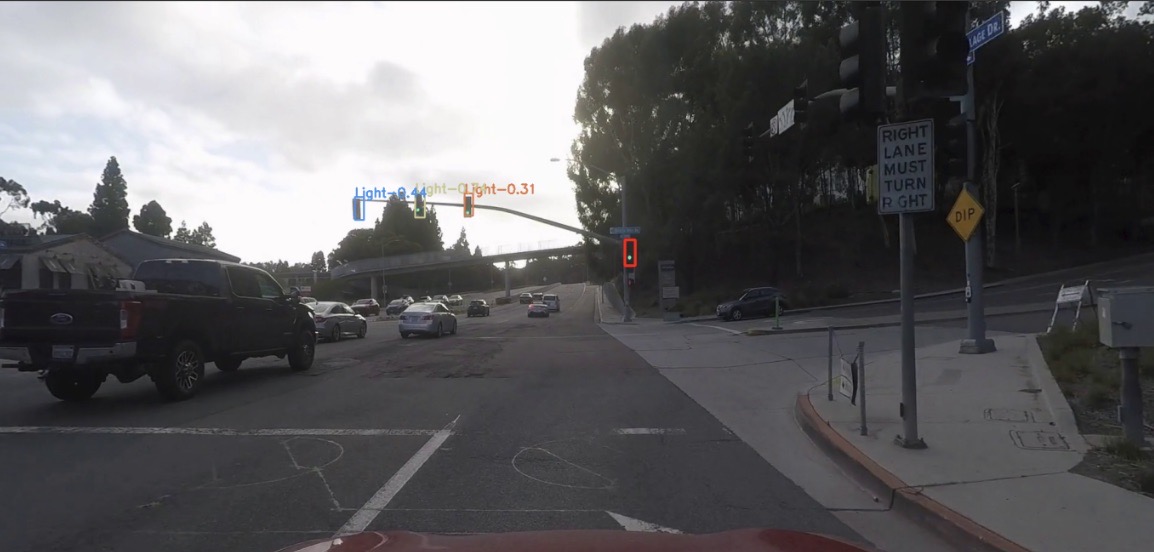}
    \includegraphics[width=.48\textwidth]{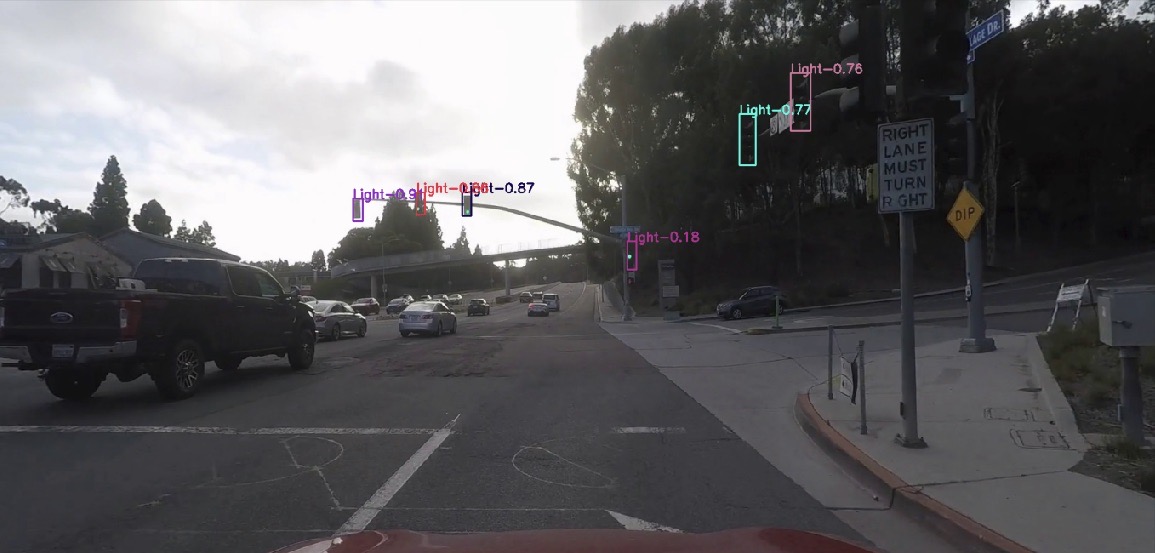}
    \label{fig:qual}
    \caption{Qualitative results of training without salience-sensitive loss (left) and with salience-sensitive loss (right). In each example, a red border is given to signs which are ``missed" by the detector. In these examples, the missed sign is salient and critical to the car's decisions. These signs are sometimes further or smaller than other signs in the scene, making salience an important part of the training process since the ``easiest" signs may not be the most important to the vehicle's planning.}

\end{figure*}

\section{Concluding Remarks}

In autonomous vehicle planning, it is not always the nearest or largest objects which provide the most critical information. Training object detectors with salience-aware methods are critical for ensuring that objects critical to the driver's decisions are emphasized in detection models. Using a transformer like Deformable DETR is a natural choice for this problem since transformers learn what portions of an image it should pay attention to. We have shown the effectiveness of salience-sensitive loss in guiding Deformable DETR toward more accurate object detection for a US traffic lights dataset. To further improve the model performance, we aim to annotate more traffic light examples to expand the LAVA Salient Signs dataset. Further research directions into salience annotation and salience-sensitive classification may continue to improve scene understanding, continuing towards an overall goal of robustly safe autonomous navigation through intersections.  

\section{Acknowledgements}
We are grateful to LISA research sponsors, especially Dr. Suchitra Sathyanarayana, Ninad Kulkarni, and Jeremy Feltracco of AWS Machine Learning Solutions Laboratory for sharing AWS Sagemaker GroundTruth 2 and LISA colleagues for their valuable assistance in creating the novel LAVA dataset and its derivatives.

\bibliography{biblio.bib}
\bibliographystyle{IEEEtran.bst}

\end{document}